\documentclass[10pt,twocolumn,letterpaper]{article}

\usepackage{iccv}
\usepackage{times}
\usepackage{epsfig}
\usepackage{graphicx}
\usepackage{amsmath}
\usepackage{amssymb}

\usepackage{multirow}
\usepackage{array}
\usepackage{float}
\usepackage{algorithm}

\usepackage[pagebackref=true,breaklinks=true,letterpaper=true,colorlinks,bookmarks=false]{hyperref}

\iccvfinalcopy

\renewcommand{\paragraph}[1]{\noindent \textbf{#1}. }

\newcommand{\argmax}{\operatornamewithlimits{argmax}}

\begin{document}
\title{Learning to track for spatio-temporal action localization}

\author{Philippe Weinzaepfel$^{a}$ \qquad Zaid Harchaoui$^{a,b}$ \qquad Cordelia Schmid$^{a}$ \\
$^{a}$ Inria\thanks{LEAR team, Inria Grenoble Rhone-Alpes, Laboratoire Jean Kuntzmann, CNRS, Univ. Grenoble Alpes, France.} \qquad \qquad $^{b}$ NYU \\
{\tt\small firstname.lastname@inria.fr}}

\maketitle

\begin{abstract}
     
We propose an effective approach for spatio-temporal action localization in realistic videos. 
The approach first detects proposals at the frame-level and scores them with a combination
of static and motion CNN features. It then tracks high-scoring proposals throughout the video using a tracking-by-detection approach. 
Our tracker relies simultaneously on instance-level and class-level detectors.
The tracks are scored using a spatio-temporal motion histogram, a descriptor at the track level, in combination with the CNN features. 
Finally, we perform temporal localization of the action  using a sliding-window approach at the track level. 
We present experimental results for spatio-temporal localization on the UCF-Sports, J-HMDB and UCF-101 
action localization datasets, where our approach outperforms 
the state of the art with a margin of 15\%, 7\% and 12\% respectively in mAP.

\end{abstract}

\section{Introduction}
\label{sec:intro}

Recent work on action recognition mostly focuses on the problem of
action classification~\cite{sports1m,PoppeSurvey,densetraj}. The goal is to assign a category label to 
a video, in most cases cropped to the extent of the action. In a long
stream of video, an  action may have varying temporal
extent. Furthermore, the action is also spatially localized.
Yet, detecting an action in space and time 
remains a challenging task which received little attention so far.

Some previous works address related issues by putting emphasis either on the spatial or on the temporal localization. 
Action recognition and localization in still images~\cite{DelaitreSL11} is an extreme example along the first line, 
where local detectors are trained \eg with HOG features and localize \emph{spatially} the person and/or the object. 
On the other extreme, recent work on action recognition and localization from videos~\cite{actoms,THUMOS14,oneataTemp}
perform \emph{temporal localization}, for which dense motion features such as dense trajectories~\cite{densetraj} proved effective.

Several recent works address spatial and temporal localization
jointly. They resort to figure-centric models~\cite{LanWM11,prest:hal-00720847,klaser:inria-00514845}, discriminative parts~\cite{STsegments} or proposals~\cite{JainCVPR2014,Yu_2015_CVPR,fat}.
Proposals are obtained by hierarchical merging of supervoxels~\cite{JainCVPR2014}, by maximizing an actionness score~\cite{Yu_2015_CVPR} or by relying on selective search regions and CNN features~\cite{fat}. 

The main challenge in spatio-temporal localization is to accommodate the uncertainty of per-frame spatial localization and the temporal consistency. 
If the spatial localization performed independently on each frame is
too selective and at the same time uncertain, then enforcing the
temporal consistency of the localization may fail.
Here we use proposals to obtain a set of per frame spatial proposals and enforce temporal consistency based on a tracker, that simultaneously relies on instance-level and class-level detectors. 

Our approach starts from frame-level proposals extracted with 
a high-recall proposal algorithm~\cite{edgeboxes}. Proposals  
are scored using CNN descriptors based on appearance and motion information~\cite{fat}.
To ensure the temporal consistency, we propose to track them with a tracking-by-detection approach combining an instance-level and class-level detector.
We then score the tracks with the CNN features as well as a spatio-temporal motion histogram descriptor, which captures the dynamics of an action.
At this stage, the tracks are localized in space, but the temporal localization needs to be determined. 
Temporal localization is performed using a multi-scale sliding-window approach at the track level.

In summary, this paper introduces an approach for spatio-temporal localization, 
by learning to track, with state-of-the-art experimental results on UCF-Sports, 
J-HMDB and UCF-101. A spatio-temporal local descriptor
allows to single out more relevant tracks and temporally localize the
action at the track level.

This paper is organized as follows. In Sec.~\ref{sec:related},
we review related work on action localization. 
We then present an overview of our approach in Sec.~\ref{sec:overview} and give the details in Sec.~\ref{sec:method}.
Finally, Sec.~\ref{sec:xp} presents  experimental results.

\section{Related work}
\label{sec:related}

Most approaches for action recognition focus on action classification~\cite{AggarwalActivity,PoppeSurvey}. 
Feature representations such as bag-of-words on space-time descriptors have shown excellent results~\cite{hog3d,LMSR08,densetraj}.
In particular, Wang \etal~\cite{wang:hal-01145834} achieve state-of-the-art performance using Fisher Vectors and dense trajectories with motion stabilization.
Driven by the success of Convolutional Neural Networks (CNNs) for many recognition tasks (image classification~\cite{Krizhevsky}, object detection~\cite{rcnn}, etc.) feature representations output by CNNs have been extended to videos. Such approaches use
3D convolutions on a stack of frames~\cite{ji3dconv,sports1m,c3d}, apply recurrent neural network on per-frame features~\cite{lrcn},  or
process images and optical flows in two separate streams~\cite{2streamsCNN}.
CNN representations now achieve comparable results to space-time descriptors.

For temporal localization of actions, most state-of-the-art approaches are based on a sliding window~\cite{actoms,wang:hal-01145834}.
To speed-up the localization, Oneata \etal~\cite{oneataTemp} proposed an approximately normalized Fisher Vector, 
allowing to replace the sliding window by a more
efficient branch-and-bound search.
The sliding-window paradigm is also common for spatio-temporal action localization.
For instance, Tian \etal~\cite{SDPM} extend the deformable part models, introduced in~\cite{dpm} for object detection in 2D images, 
to 3D images by using HOG3D descriptors~\cite{hog3d} and employ a sliding window approach, in scale, space and time.
Wang \etal~\cite{WangQT14} first use a temporal sliding window and then model the relations between dynamic-poselets.
Laptev and Perez~\cite{LaptevP07} perform a sliding window on cuboids, thus restricting the action to have a fixed spatial extent across frames.

Another category of action localization approaches uses a figure-centric model.
Lan \etal~\cite{LanWM11} learn a spatio-temporal model for an action using a figure-centric visual word representation,
where the location of the subject is treated as a latent variable and is inferred jointly with the action label. 
Prest \etal~\cite{prest:hal-00720847} propose to detect humans and objects and then model their interaction.
Humans detectors were also used by Kl\"{a}ser \etal~\cite{klaser:inria-00514845} for action localization.
The detected humans are then tracked across frames using optical flow and the track is classified using HOG-3D~\cite{hog3d}.
Our approach also relies on tracking, but is more robust to appearance and pose changes by using a tracking-by-detection approach~\cite{struck,tld},
in combination with a class-specific detector.
In addition, we classify the tracks using per-frame CNN features and spatio-temporal features.

Some other methods are based on the generation of action proposals~\cite{JainCVPR2014,Yu_2015_CVPR}. Yu and Yuan~\cite{Yu_2015_CVPR} 
compute an actionness score and then use a greedy method 
to generate proposals. Jain~\etal~\cite{JainCVPR2014} propose a method based on merging a hierarchy of supervoxels. 
Ma~\etal~\cite{STsegments} leverage a hierarchy of discriminative parts to represent and localize an action.
The extension of structured output learning from object detection to action localization was proposed by Tran and Yuan~\cite{NIPS2012_4794}.
Recently, Gkioxari and Malik~\cite{fat} proposed to use object proposals for action localization. 
Object proposals from SelectiveSearch~\cite{selectivesearch} are detected in each frame, scored using features from a two-streams CNN architecture, and linked across the video.
Our approach is more robust since we do not force detections to pass through proposals at every frame. 
Moreover, we combine the per-frame CNN features with descriptors extracted at a spatio-temporal level to capture the dynamics of the actions.

\section{Overview of the approach}
\label{sec:overview}

\begin{figure*}
 \centering
 \includegraphics[width=\linewidth]{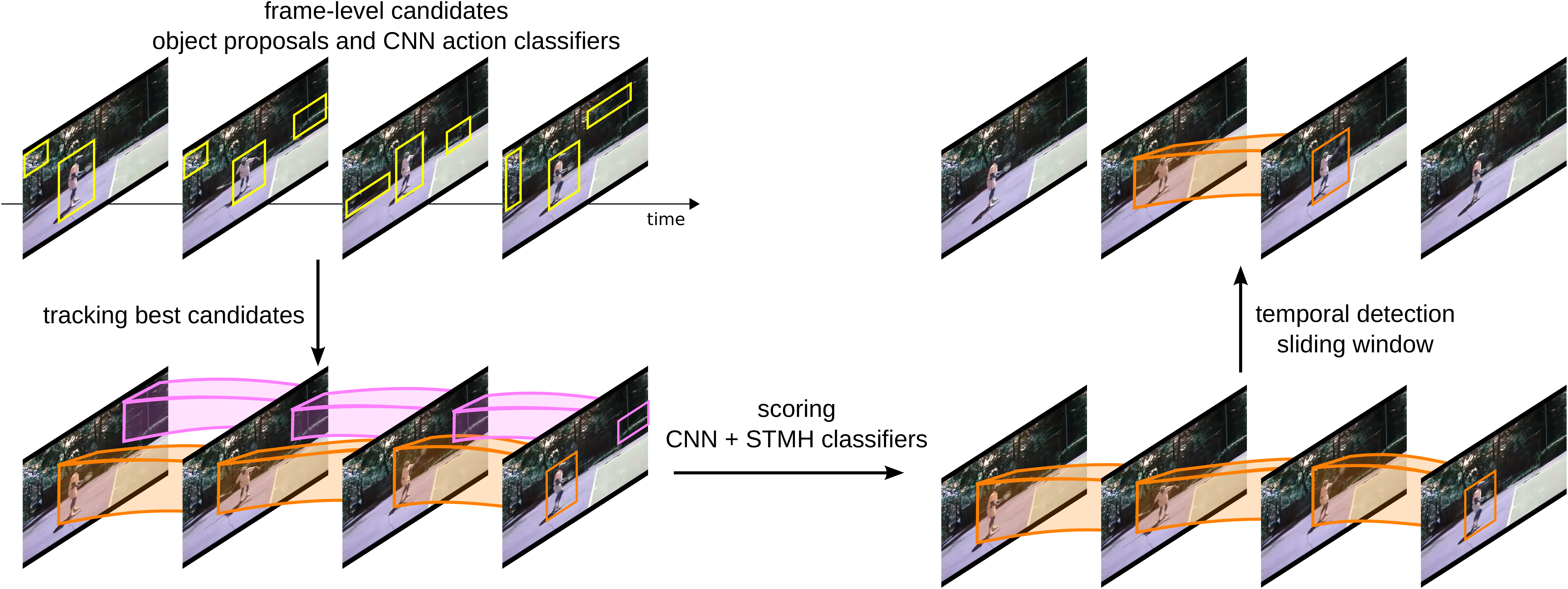}
 \caption{Overview of our action localization approach. We detect frame-level object
   proposals and score them with CNN action classifiers. The best
   candidates, in term of scores, are tracked throughout the video. We then score the tracks with CNN and spatio-temporal motion histogram (STMH) classifiers. 
 Finally, we perform a temporal sliding window for detecting the temporal extent of the action.
 }
 \label{fig:outline}
 
 \vspace{-0.3cm}
 
\end{figure*}

Our approach for spatio-temporal action localization consists of four stages, see
Figure~\ref{fig:outline}. We now briefly present them and 
then provide a detailed description in Section~\ref{sec:method}.    

\noindent {\bf Extracting and scoring frame-level proposals.} Our
method extracts a set of candidate regions at the frame level. We use
EdgeBoxes~\cite{edgeboxes}, as they obtain a high recall even when considering relatively few 
proposals~\cite{Hosang2015arXiv}. Each proposal is represented with
CNN features~\cite{fat}. These CNN features leverage both static and
motion information and   
are trained to discriminate the actions against background regions.
This is capital since most of the proposals do not contain any action.
For each class, a hard negative mining procedure is performed in order
to train an action-specific classifier. Given a test video,
frame-level candidates are scored with these action-specific
classifiers.  

\noindent {\bf Tracking best candidates.} Given the 
frame-level candidates of a video,  we select the highest scoring ones per class
and track them throughout the video. Our tracking method is based on a
standard tracking-by-detection approach leveraging an instance-level detector
as well as a class-level classifier. The detector is based on the same
CNN features as the first stage. We perform the tracking multiple
times for each action, starting from the proposal with the highest
score that do not overlap with previous computed tracks.  

\noindent {\bf Scoring tracks.}
The CNN features only contain information extracted at the frame level. 
Consequently, they are not able to capture the dynamics of an action across multiple frames.
Thus, we introduce a spatio-temporal motion histogram (STMH). It is
inspired by the success of dense trajectory
descriptors~\cite{densetraj}. Given a fixed-length chunk from a track,
we divide it into spatio-temporal cells and compute an histogram of
gradient, optical flow and motion boundaries in each cell. A 
hard-negative mining is employed to learn a classifier for each
class. The final score is obtained by combining CNN and STMH 
classifiers.  

\noindent {\bf Temporal localization.} To detect the temporal extent
of an action, we use a multi-scale sliding window approach over tracks.  At test time, we rely 
on temporal windows of different
lengths that we slide with a stride of 10 frames over the tracks. We
score each temporal window according to CNN features, STMH descriptor and a duration prior learned on the training set.
For each track, we then select the window with the highest score.

\section{Detailed description of the approach}
\label{sec:method}

In this section, we detail the four stages of our action localization approach.
Given a video of $T$ frames $\{ I_t \}_{t=1..T}$ and a class $c \in \mathcal{C}$ ($\mathcal{C}$ being the set of classes), 
the task consists in detecting if the action $c$ appears in the video and if yes, when and where.
In other words, the approach outputs a set of regions $\{ R_t \}_{t=t_b..t_e}$ with $t_b$ (resp. $t_e$) 
the beginning (resp. end) of the predicted temporal extent of the action $c$ and $R_t$ the detected region in frame $I_t$.

\subsection{Frame-level proposals with CNN classifiers}
\label{subsec:proposal}

\paragraph{Frame-level proposals}
State-of-the-art methods~\cite{rcnn} for object localization replace the sliding-window paradigm used in the past decade by object proposals.
Instead of scanning the image at every location, at several scales, 
object proposals allow to significantly reduce the number of candidate regions, 
and narrow down the set to regions that are most likely to contain an object. 
For every frame, we extract EdgeBoxes~\cite{edgeboxes} using the online code and keep the best 256 proposals according to the EdgeBox score.
We denote by $\mathcal{P}_t$ the set of object proposals for a frame $I_t$.
In Section~\ref{subsec:tracking}, we introduce a tracking approach that makes our method robust to missing proposals.

\paragraph{CNN features}
Recent work on action recognition~\cite{2streamsCNN} and localization~\cite{fat} have demonstrated the benefit of CNN feature
representations, applied
separately on images and optical flows. We use the same set of
CNN features as in~\cite{fat}.

\begin{figure}
 \centering
 \includegraphics[width=\linewidth]{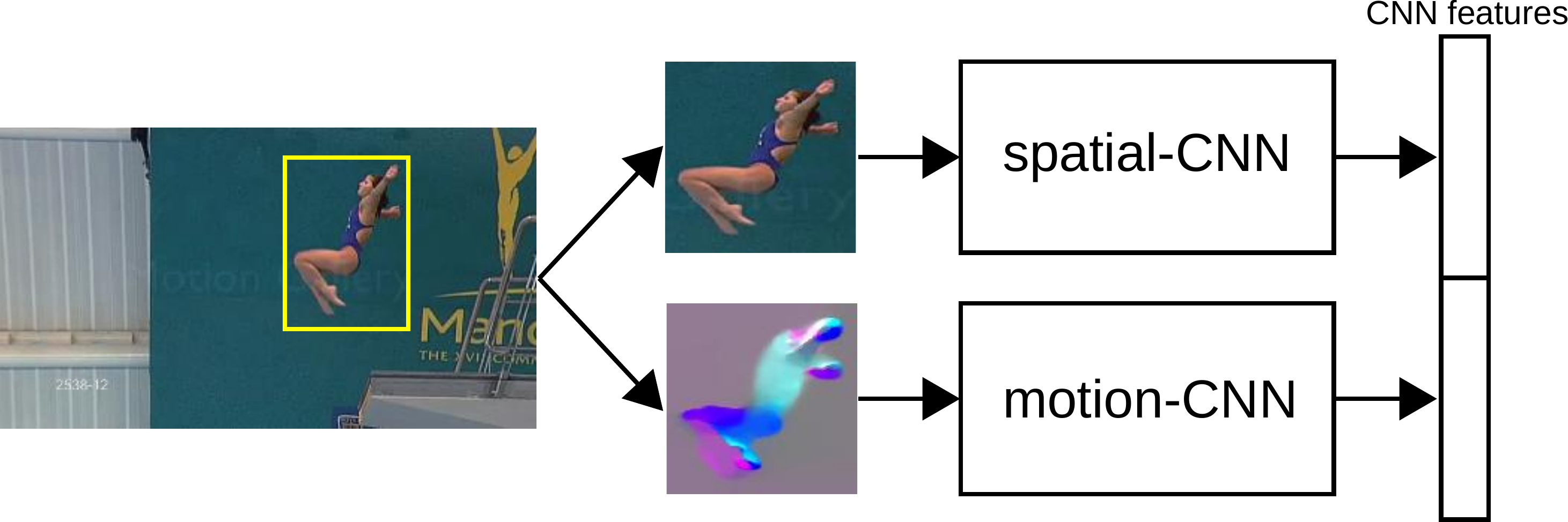}
 \caption{Illustration of CNN features for a region $R$. The CNN
   features are the concatenation of the fc7 layer from the
   spatial-CNN and motion-CNN, \ie, a 2x4096 dimensional descriptor. }
 \label{fig:cnnfeat}
 
 \vspace{-0.5cm}
 
\end{figure}

Given a region resized to $227 \times 227$ pixels, a spatial-CNN operates on RGB channels and captures the static appearance of the actor and the scene, 
while a motion-CNN takes as input optical flow and captures motion pattern.
The optical flow signal is transformed into a 3-dimensional image by stacking the x-component, the y-component and the magnitude of the flow. Each image is then multiplied by $16$
and converted to the closest integer between 0 and 255.
In practice, optical flow is estimated using the online code 
from Brox~\etal~\cite{Bro04a}.
For a region $R$, the CNN features we use are the concatenation of the
fc7 layer (4096 dimensions) from the spatial-CNN and motion-CNN, see
Figure~\ref{fig:cnnfeat}. 

\paragraph{CNN training}
We use the same architecture and training procedure as~\cite{fat}. We
give a brief presentation below and refer to their work for more details.
The architecture is the same for both networks with 5 convolution layers interleaved by pooling and normalization, and then 3 fully connected layers interleaved with dropout. 
The last fully connected layer (fc8) has $\vert \mathcal{C} \vert +1$ outputs, one per class and an additional output for the background.
Similar to~\cite{rcnn}, during training, the proposals that overlap more than $50\%$ with the ground-truth are considered as positives, the others as background.
Regions are resized to fit the network size ($227 \times 227$) and randomly flipped. 
The spatial-CNN is initialized with a model trained on full images from ImageNet and fine-tuned for object detection on Pascal VOC 2012~\cite{rcnn}. 
For the motion-CNN, initialization weights are trained for the task of
action recognition on the UCF-101 dataset~\cite{ucf101} with full frames of the training set of split 1. 
We then fine-tune the networks with back-propagation using Caffe
~\cite{caffe} on the proposal regions for each dataset.
Each batch contains $25\%$ of non-background regions.

\paragraph{Action classifiers}
For each action class $c \in \mathcal{C}$, we train a linear SVM using hard negative mining.
The positives are given by the ground-truth annotations and negatives
by all proposals whose overlap with a ground-truth region is below
$30\%$. 
At test time, we denote by $S_{\text{CNN}}(c,R)$  the score of a
region $R$ for the action class $c$ given by the trained classifier.
This yields a confidence score for the region $R$ and an  action
class $c$.

\subsection{Tracking}
\label{subsec:tracking}

The second stage consists in tracking the best proposals over the video.
We use a tracking-by-detection approach that leverages instance-level and class-level detectors.
Let $R$ be a region in frame $I_\tau$ for the class $c$ to be tracked. 
As a result, the tracking stage will output a track $\mathcal{T}_c = \{ R_t \}_{t=1..T}$. 
The track provides a candidate localization for the action $c$. 
We first present how the tracker is initialized.
Then, we detail the tracking procedure. 
Finally, we explain the selection of the regions to track.

\paragraph{Initialization}
Given a region $R$ to be tracked in frame $I_\tau$, the first step is to refine the position and size of the region by 
performing a sliding-window search both in scale and space in the neighborhood of $R$.
Let $\mathcal{N}(R)$ be the set of windows scanned with a sliding window around the region $R$.
The best region according to the action-level classifier is selected: $R_\tau = \argmax_{r \in \mathcal{N}(R)} S_{\text{CNN}}(c,r)$.
The sliding-window procedure using CNN features can be performed efficiently~\cite{Giusti2013a,overfeat}.

Given the refined region, we train an instance-level detector using a linear SVM.
The set of negatives comprises the instances extracted from boxes whose
overlap with the original region is less than $10\%$. The boxes are restricted to regions in $\mathcal{P}_\tau$, \ie, the proposals in frame $\tau$. 
The set of positives is restricted to the refined region $R_\tau$. This strategy is consistent with current tracking-by-detection approaches~\cite{huatrack}. 
Denote by $S_\text{inst}(R)$ the score of the region $R$ with the instance-level classifier.
We now present how the tracking proceeds over the video. We first do a forward pass from frame $I_\tau$ to the last frame $I_T$, 
and then a backward pass from frame $I_\tau$ to the first frame.

\begin{algorithm}
\small

\textbf{Input:} a region $R$ in frame $I_\tau$ to track, a class $c$

\textbf{Output:} a track $\mathcal{T}_c = \{ R_t \}_{t=1..T}$

$R_\tau \leftarrow \argmax_{r \in \mathcal{N}({R}) } S_{\text{CNN}}(c,r)$ 

$\textsf{Pos} \leftarrow \{ R_\tau \}$ 

$\textsf{Neg} \leftarrow \{ r \in \mathcal{P}_\tau \vert \: \text{IoU}(r,R_\tau)<0.1 \}$  

\textbf{For} $i = \tau+1~...~T \text{ and } \tau-1~...~1$: 

$\quad$ Learn instance-level classifier from $\textsf{Pos}$ and $\textsf{Neg} $ 

$\quad$ $R_i \leftarrow \argmax_{r \in \mathcal{N}(R'_{i})} ( S_{\text{CNN}}(c,r) + S_{\text{inst}}(r) )$

$\quad$ $\textsf{Neg} \leftarrow \textsf{Neg} \cup \{ r \in \mathcal{P}_\tau \vert \: \text{IoU}(r,R_i)<0.1 \}$ 

$\quad$ $\textsf{Neg} \leftarrow \{ r \in \textsf{Neg} \vert S_{\text{inst}}(r) \geq -1 \}$ ~~~(restrict to hard negatives)

$\quad$ $\textsf{Pos} \leftarrow \textsf{Pos} \cup \{R_i\}$ 

\normalsize
\caption{Tracking}
\label{alg:track} 
\end{algorithm}

\paragraph{Update}
Given a tracked region $R_t$ in frame $I_t$, we now want to find the most likely location in frame $I_{t+1}$.
We first map the region $R_t$ into $R'_{t+1}$, by shifting the region with the median of the flow 
between frame $I_t$ and $I_{t+1}$ inside the region $R_t$. We then select the best region in the neighborhood of $R'_{t+1}$  using a sliding window that leverages both class-level and instance-level classifiers:  
\begin{equation}
R_{t+1} = \argmax_{r \in \mathcal{N}(R'_{t+1})} S_{\text{inst}}(r) + S_{\text{CNN}}(c,r) ~~.
\end{equation}

In addition, we update the instance-level classifier by adding $R_{t+1}$ as a positive exemplar and proposals $\mathcal{P}_{t+1}$ from frame $I_{t+1}$
that do not overlap with this region as negatives.
Note that at each classifier update, we restrict the set of negatives to the hard negatives.

The tracking algorithm is summarized in Algorithm~\ref{alg:track}.
By combining instance-level and class-level information, our tracker is robust to significant changes in appearance and occlusion.
Note that category-specific detectors were previously used in other contexts, such as face~\cite{facetld} or people~\cite{doi:10.5244/C.24.55} tracking.
We demonstrate the benefit of such detectors in our experiments in Section~\ref{sec:xp}.

\paragraph{Proposals selection}
We now present how we chose the proposals to track.
We first select the subset of classes for which the tracking is performed.
To this end, we assign a score to the video for each class $c \in \mathcal{C}$ and keep the top-5.
The score for a class $c$ is defined as $\max_{r \in \mathcal{P}_t, t=1..T} S_{\text{CNN}}(c,r)$, \ie, we keep the maximum score for $c$ over all proposals of the video. 

When generating tracks for the class $c$, we first select the proposal with the highest score over the entire video.
We run the tracker starting from this region and obtain a first track.
We then perform the tracking iteratively, starting a new track from the best proposal that does not overlap with any previous track from the same class.
In practice, we compute 2 tracks for each selected class.

\subsection{Track descriptor}
\label{subsec:descriptor}

So far, we have only used features extracted on individual frames. Clearly, this does not capture the dynamics of the action over time.
To overcome this issue, we introduce a spatio-temporal motion histogram (STMH) feature.

\begin{figure}
 \centering
 \includegraphics[width=0.8\linewidth]{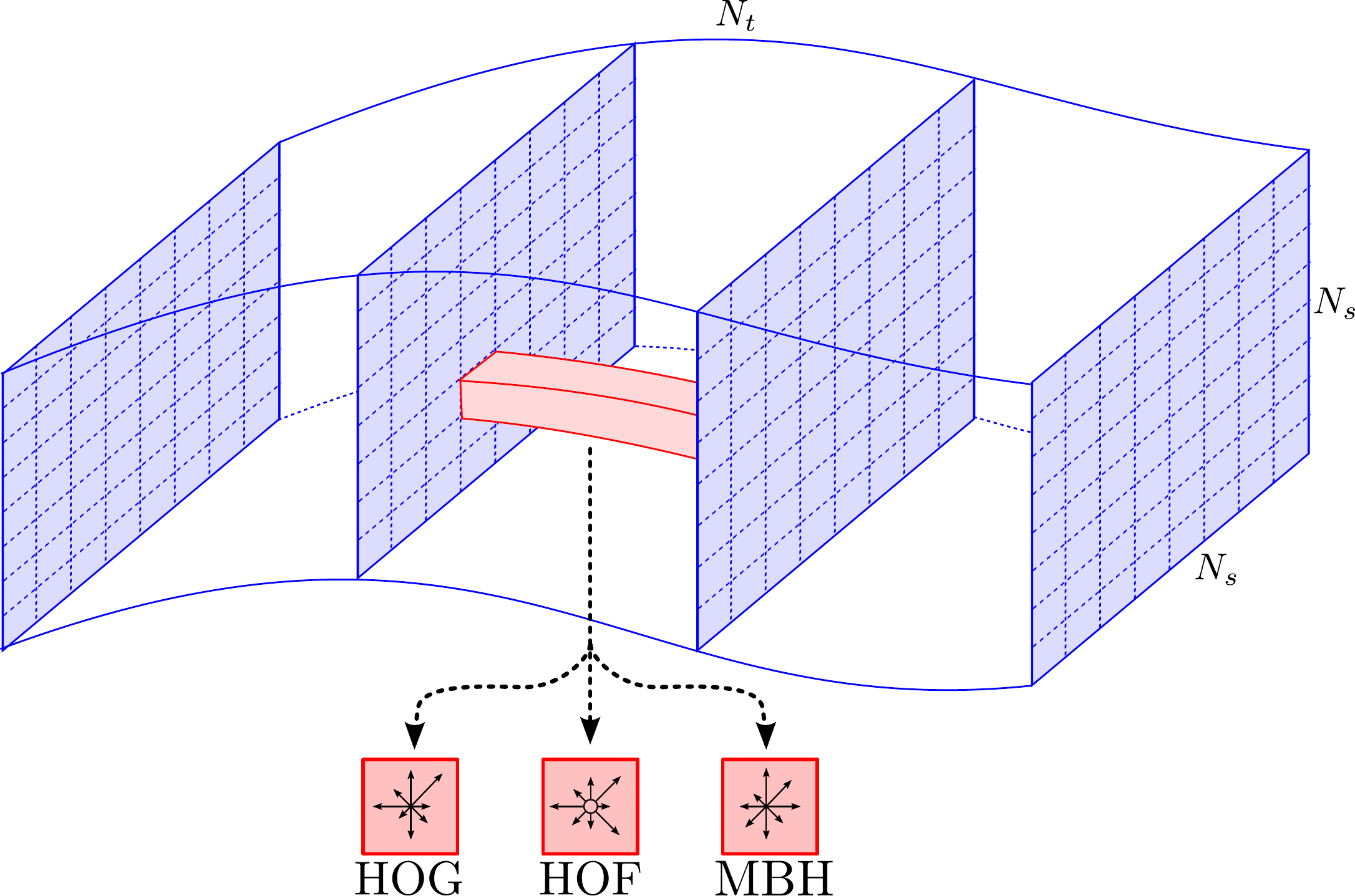}
 \caption{Illustration of STMH. A chunk is split into spatio-temporal cells for which an histogram of gradient, optical flow and motion boundaries is computed.}
 \label{fig:desc}
 
 \vspace{-0.2cm}
 
\end{figure}

\paragraph{The STMH descriptor}
Similar to Wang \etal~\cite{densetraj}, we rely on histograms of gradient and motion extracted in spatio-temporal cells.
Given a track $\mathcal{T}_c = \{ R_t \}_{t=1..T}$, it is divided into temporal chunks of $L=15$ frames, with a chunk starting every $5$ frames.
Each chunk is then divided into $N_t$ temporal cells, and each region $R_t$ into $N_s \times N_s$ spatial cells, as shown in Figure~\ref{fig:desc}.
For each spatio-temporal cell, we perform a quantization of the per-pixel image gradient into an histogram of gradients (HOG) with 8 orientations. The histogram is then normalized with the L2-norm.
Similarly, we compute HOF, MBHx and MBHy by replacing the image gradient by the optical flow and the gradient of its x and y components.
For HOF, a bin for an almost zero value is added, with a threshold at $0.04$.
In practice, we use $3$ temporal cells and $8 \times 8$ spatial cells, resulting in $3 \times 8 \times 8 \times (8+9+8+8)=6336$ dimensions.
Note that we use more spatial cells than~\cite{densetraj}, as our regions are on average significantly larger than the $32 \times 32$ patch they use.

\paragraph{Fusion}
For each action, we train a linear SVM using hard negative mining.
The set of positives is given by features extracted along the ground-truth annotations, 
while the negatives are given by cuboids (spatially and temporally) centered at the proposals that do not overlap with 
the ground-truth. 
Let $S_\text{desc}(c, \mathcal{T})$  be the average of the scores for all the chunks of length $L$ from the track $\mathcal{T}$ for the action $c$.

Given a track $\mathcal{T} = \{R_t\}_{t=1..T}$, we score it by summing the scores from the CNN averaged over all frames, and the scores from the descriptors averaged over chunks: 
\begin{equation}
S(\mathcal{T}) = \sigma \Big( S_{\text{desc}}(c, \mathcal{T}) \Big) + \sigma \Big( \sum_{t=1}^{T} S_{\text{CNN}}(c,R_t) \Big) \, ,
\label{eqn:score}
\end{equation}
where $\sigma (x) = 1/(1+e^{-x})$.
We summarize the resulting approach for spatio-temporal detection in Algorithm~\ref{alg:detect}.

\subsection{Temporal localization}
Similar to the winning approach in the temporal action detection track of the Thumos 2014 challenge~\cite{learthumos14}, we use a sliding-window strategy for temporal localization. However, we apply the sliding window directly on each track $\mathcal{T}$, 
while ~\cite{learthumos14} used features extracted for the full frames. 
The window length takes values of 20, 30, 40, 50, 60, 70, 80, 90, 100, 150, 300, 450 and 600 frames. The sliding window has a stride of 10 frames.
For each action $c$, we learn the frequency of its durations on the training set.
We score each window using the score described above based on CNNs features and STMH, normalized with a sigmoid, and multiply it with the per-class duration prior.
For each track, we keep the top-scoring window as spatio-temporal detection.

\begin{algorithm}[t]
\small

\textbf{Input:} a test video $\{I_t\}_{t=1...T}$

\textbf{Output:} a list of detections $(c, \mathcal{T}, \text{score})$

\textbf{For} $t=1..T$ 

$\quad$ $\mathcal{P}_t = \text{EdgeBoxes}(I_t)$ 

$\quad$ \textbf{For} $r \in \mathcal{P}_t$ 

$\quad$ $\quad$ Compute $S_{\text{CNN}}(c, r)$ 

$\mathcal{C}' \leftarrow$ class selection (see Sec.~\ref{subsec:tracking}) 

$\text{Detections} \leftarrow [~]$

\textbf{For} $c \in \mathcal{C}'$ 

$\quad$ \textbf{For} $i = 1...\text{ntracks}$ {\scriptsize (we generate ntracks=2 tracks per label, see Sec.~\ref{subsec:tracking})}

$\quad$ $\quad$ $R,\tau \leftarrow \argmax_{r\in \mathcal{P}_t, t=1..T} S_{\text{CNN}}(c,r)$ 

$\quad$ $\quad$ $\quad$ $\quad${\footnotesize (proposal to track without overlap with previous tracks)}

$\quad$ $\quad$ $\mathcal{T} \leftarrow \text{Tracking}(R,I_\tau,c)$ ~~~~~(Algorithm~\ref{alg:track})

$\quad$ $\quad$ $\text{score} \leftarrow \sigma ( S_{\text{STMH}}(c, \mathcal{T}) ) + \sigma( \sum_{R_t \in \mathcal{T}} S_{\text{CNN}}(c, R_t) )$ (Eq.~\ref{eqn:score})

$\quad$ $\quad$ $\text{Detections} \leftarrow \text{Detections} \cup \{ (c, \mathcal{T}, \text{score}) \}$ 

\normalsize
\caption{Spatio-temporal detection in a test video}
\label{alg:detect} 
\end{algorithm}

\section{Experimental results}
\label{sec:xp}

In this section, we first present the datasets and the evaluation protocol.
We then study the impact of both the tracking and the class selection, and provide a parametric study of STMH.
Finally, we show that our approach outperforms the state of the art for spatio-temporal action localization.

\subsection{Datasets and evaluation}

In our experiments, we use three datasets: UCF-Sports, J-HMDB and UCF-101. 

\noindent {\bf UCF-Sports~\cite{ucfsports}.} The dataset contains 150 short videos
of sports broadcasts from 10 actions classes: diving, golf
swinging, kicking, lifting, horse riding, running, skating,
swinging on the pommel horse and on the floor, swinging at the high bar and walking. 
Videos are truncated to the action and bounding boxes annotations are provided for all frames. 
We use the standard training and test split defined in~\cite{LanWM11}.

\noindent {\bf J-HMDB~\cite{jhmdb}.} The dataset is a subset of the HMDB dataset~\cite{hmdb}.
It consists of 928 videos for 21 different actions such as brush hair,
swing baseball or jump. Video clips are restricted to the duration of the action.
Each clip contains between 15 and 40 frames. 
Human silhouettes are annotated for all frames. 
Thus, the dataset can be used for evaluating action localization.
There are 3 train/test splits and evaluation averages the results over the three splits. 

\begin{figure}
 \centering
 \includegraphics[width=0.8\linewidth]{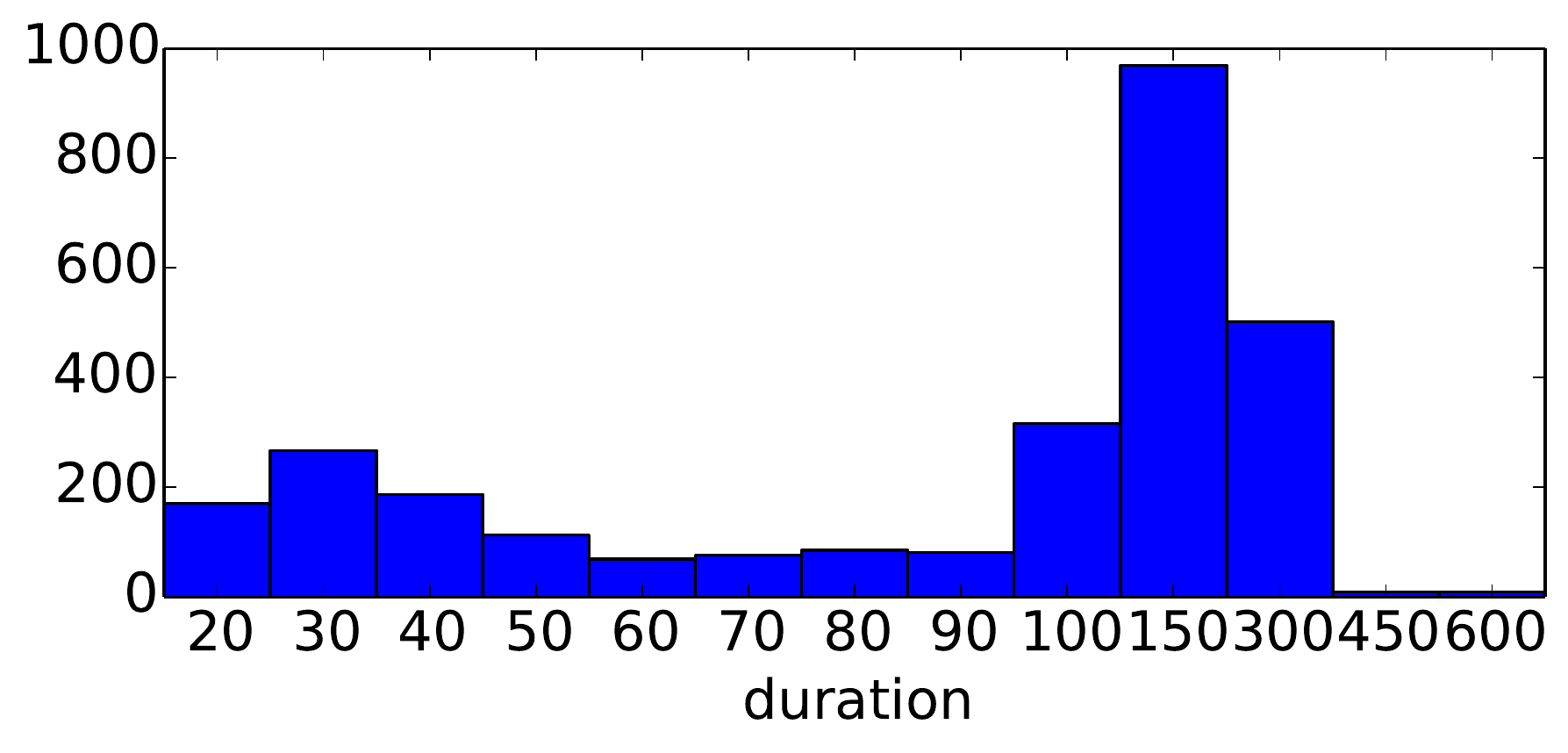}
 \caption{Histogram of action durations for the 24 classes
   with spatio-temporal annotations in the UCF-101 dataset (training set).}
 \label{fig:durations}
 
 \vspace{-0.1cm}
 
\end{figure}

\noindent {\bf UCF-101~\cite{ucf101}.}
The dataset is dedicated to action classification with more than 13000 videos and 101 classes.
For a subset of 24 labels, the spatio-temporal extents of the actions are annotated.
All experiments are performed on the first split only. 
In contrast to UCF-Sports and J-HMDB where the videos are truncated to the action, UCF-101 videos are longer and the localization is both spatial and temporal.
Figure~\ref{fig:durations} shows a histogram of the action durations
in the training set, averaged over all 24 classes. 
Some of the actions are long, such as `soccer juggling' or `ice dancing', whereas others last only few frames, \eg `tennis swing' or `basketball dunk'.

\noindent {\bf Evaluation metrics.}
A detection is considered ``correct'' if the intersection over union
(IoU) with the ground-truth is above a threshold $\delta$.
The IoU between two tracks is defined as the IoU over the temporal domain, 
multiplied by the average of the IoU between boxes averaged over all overlapping frames.
Duplicate detections are considered as ``incorrect''.
By default, the reported metric is the mean Average Precision at threshold $\delta = 50\%$ for spatial localization (UCF-Sports and J-HMDB) and $\delta = 20\%$ for 
spatio-temporal localization (UCF-101).
When comparing to the state of the art on UCF-Sports, we also use ROC curves and report the Area Under the Curve (AUC) as done by previous work.
Note that this metric is impacted by the set of negatives detections
and, thus, may not be suited for a detection task~\cite{pascal11}.
Indeed, if one adds many easy negatives, \ie, negatives that are ranked
after all positives, the AUC increases while the mAP remains the same. 

\subsection{Impact of the tracker}

\begin{table}
\resizebox{\linewidth}{!}{
 \begin{tabular}{|c||c|c||c|c|}
 \hline
 \multirow{2}{*}{Detectors in the tracker} & \multicolumn{2}{c||}{recall-track}  & \multicolumn{2}{c|}{mAP} \\
 & UCF-Sports & J-HMDB & UCF-Sports & J-HMDB \\
 \hline
            instance-level + class-level & \textbf{98.75\%} &  91.74\% & \textbf{90.50\%} &  \textbf{59.74\%}  \\
\hline
                     instance-level only & 85.42\% & \textbf{94.59\%} & 74.27\% &  54.32\% \\
\hline
                        class-level only & 92.92\% & 81.28\% & 85.67\% &  53.25\% \\
\hline
 \end{tabular}
 }
 \caption{ Impact of the detectors used in the tracker. We measure if the tracks generated for the ground-truth label cover the ground-truth tracks (recall-track).
 We also measure the impact of the tracker on the final detection performance (mAP). The experiments are done on UCF-Sports and J-HMDB (split 1 only).}
 \label{tab:tracking}
 
 \vspace{-0.2cm}
 
\end{table}

The strength of our approach lies in the combination of class-specific and instance-level detectors in the tracker.
To measure the benefit of this combination, Table~\ref{tab:tracking} compares the performance when removing one of them.
`Recall-tracks' measures if at least one of the 2 generated tracks for the ground-truth action covers the ground-truth annotations (IoU $\geq 0.5$), 
\ie, it measures the recall at the track level.
We also measure the impact on the final detection performance (mAP) by running our full pipeline with each tracker.

On UCF-Sports, tracking obtained by combining the detectors leads to the highest recall.
Using the instance-level detector significantly degrades the recall by $13\%$.
This can be explained by the abrupt changes in pose and appearance for actions such as diving or swinging.
On the other hand, the instance-level detector performs well on the J-HMDB dataset, which contains more static actions

Combining instance-level and class-specific classifiers also gives the best performance in term of final detection results.
On UCF-Sports, this is mainly due to the higher recall.
On J-HMDB, we find that using the instance-level detector only leads to a better recall but the precision decreases because there are 
more tracks from an incorrect label that have a high score.

Table~\ref{tab:tracklink} compares the localization mAP on UCF-Sports when using our proposed tracker or a linking strategy as in~\cite{fat}.
We experiment with proposals from SelectiveSearch~\cite{selectivesearch} (approximately $2700$ proposals per frame) or EdgeBoxes~\cite{edgeboxes} (top-256), with or without STMH.
We can see that using EdgeBoxes instead of SelectiveSearch leads to a gain of $6\%$ when using STMH. Using a tracking strategy leads to a further gain of $7\%$,
with in addition a more refined localization, see Figure~\ref{fig:ucfsports}. This shows that the tracker is a key component to the success of our approach.

\begin{table}
 \centering
 \resizebox{\linewidth}{!}{
 \begin{tabular}{|c||c|c||c|c|}
  \hline
  & \multicolumn{2}{c||}{without STMH} & \multicolumn{2}{c|}{ with STMH} \\
   &  Linking & Tracking  &  Linking & Tracking \\
  \hline
  SelectiveSearch~\cite{selectivesearch} & 75.94\% & 83.77\% & 77.1\% & 84.9\% \\
  \hline
  EdgeBoxes-256~\cite{edgeboxes} & 79.89\% & \textbf{88.23\%} & 83.2\% & \textbf{90.5\%} \\
  \hline
 \end{tabular}
 }
 \caption{ Comparison of tracking and linking,  SelectiveSearch and EdgeBoxes-256 proposals without and with STMH on UCF-Sports (localization in mAP).}
 \label{tab:tracklink}
 
 \vspace{-0.2cm}
 
\end{table}

\subsection{Class selection}

\begin{figure}
 \centering
 \hfill
 \includegraphics[width=0.49\linewidth]{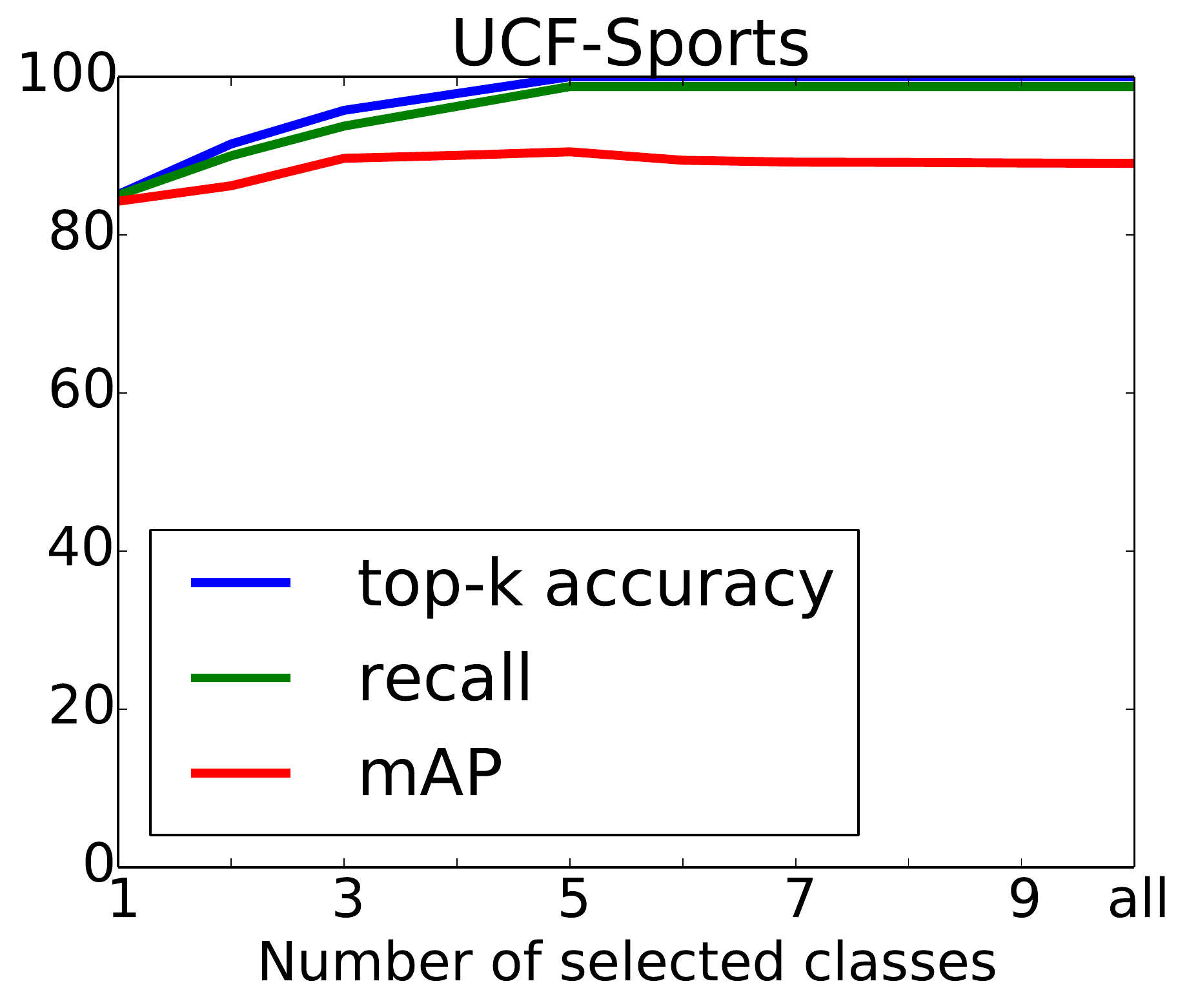}
 \hfill
 \includegraphics[width=0.49\linewidth]{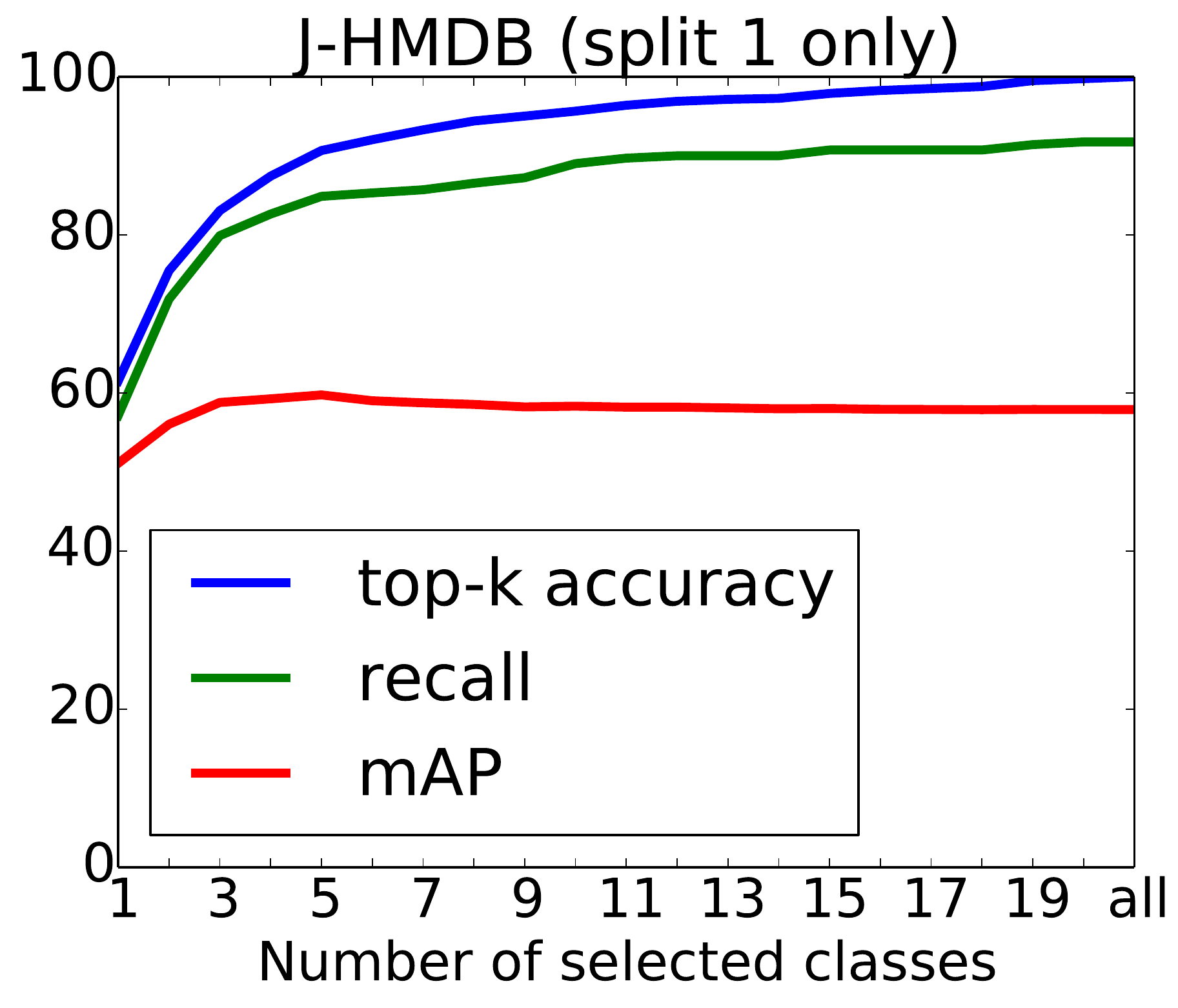}
 \hfill
 \caption{Impact of the class selection on UCF-Sports (\textit{left}) and J-HMDB (\textit{right}) datasets. In blue, top-k accuracy is shown, \ie, the percentage of cases 
 where the correct label is in the top-k classes. The recall when changing the number of selected classes is shown in green and the mAP in red.}
 \label{fig:selection}
 
  \vspace{-0.2cm}
 
\end{figure}

We now study the impact of selecting the top-5 classes based on the
maximum score over all proposals from a video for a given
class, see Section~\ref{subsec:tracking}. 
We measure the percentage of cases where the correct label is in the
top-k classes and shows the results in Figure~\ref{fig:selection} (blue curve). 
Most of the time, the correct class has the highest maximum score (around 85\% on UCF-Sports and 61\% on J-HMDB).
If we use top-5, we misclassify less than $10\%$ of the
videos on J-HMDB, and $0\%$ on UCF-Sports. 

Figure~\ref{fig:selection} also shows that recall (green) is lower than
the top-k accuracy because  the generated tracks might not have a
sufficient overlap with the ground-truth due to a failure of the tracker. 
The difference between recall and top-k accuracy is more important for large k.
This can be explained by the fact that the class-level detector performs poorly for videos where the correct label has a low rank, therefore the class-specific tracker performs poorly as well.

In addition, we display in red the evolution of the mAP on UCF-Sports and J-HMDB (split 1 only) when changing the number of selected classes.
Initially, the performance significantly increases as this corrects the cases where the correct label is top-k but not first, \ie, the recall increases.
The performance then saturates since, even in the case where a new correct label is tracked over a video, 
the final score will be low and will not have an important impact on the precision.
As a summary, selecting the top-k classes performs similar as keeping all classes while it significantly reduces the computational time.

\subsection{STMH parameters}

\begin{table}
\centering
\footnotesize
 \begin{tabular}{|c|c|c||c|c|}
 \hline
  $N_t$ & $N_s$ & dimension & UCF-Sports & J-HMDB \\
 \hline
  \multirow{4}{*}{1} &  2 &   132 &  76.07\% &  38.78\% \\
   &  4 &   528 &  80.00\% &  48.58\% \\
   &  8 &  2112 &  82.50\% &  51.71\% \\
   & 16 &  8448 &  81.67\% &  49.54\% \\
 \hline
  \multirow{4}{*}{2} &  2 &   264 &  77.98\% &  41.21\% \\
   &  4 &  1056 &  80.00\% &  49.41\% \\
   &  8 &  4224 &  87.50\% &  52.72\% \\
   & 16 & 16896 &  82.50\% &  48.89\% \\
  \hline
  \multirow{4}{*}{3} &  2 &   396 &  82.74\% &  41.38\% \\
   &  4 &  1584 &  83.33\% &  50.52\% \\
   &  8 &  6336 &  87.50\% &  \textbf{54.26\%} \\
   & 16 & 25344 &  84.17\% &  47.98\% \\
  \hline
  \multirow{4}{*}{5} &  2 &   660 &  79.64\% &  41.51\% \\
   &  4 &  2640 &  80.00\% &  50.84\% \\
   &  8 & 10560 &  \textbf{88.33\%} &  52.11\% \\
   & 16 & 42240 &  84.17\% &  47.81\% \\
 \hline
 \end{tabular}
 \normalsize
 \caption{Comparison of mean-Accuracy when classifying ground-truth tracks using STMH with different numbers of temporal ($N_t$) and spatial ($N_s$) cells.}
 \label{tab:desc}
 
 \vspace{-0.4cm}
 
\end{table}

We now study the impact of the number of temporal and spatial cells in STMH.
For evaluation, we consider the classification task and learn a linear SVM on the descriptors extracted from the ground-truth annotations of the training set.
We then predict the label on the test set, assuming the ground-truth localization is known, and report mean Accuracy.
Results are shown in Table~\ref{tab:desc}.
We can see that the best performance is obtained with $N_s=8$ spatial cells on both datasets, independently of the number of temporal cells $N_t$.
By increasing the number of cells to a higher value, \eg $16$, the descriptor becomes too specific for a class.
When using a unique temporal cell, \ie, $N_t=1$, the performance is significantly worse than for $N_t=3$.
We choose $N_s=8$ and $N_t=3$ in the remainder of the experiments.
The resulting STMH descriptor has $6,336$ dimensions.

Using the same protocol, we obtain a performance of 91.9\% for UCF-Sports and 57.99\% 
for J-HMDB using state-of-the-art improved dense
trajectories~\cite{wang:hal-01145834} with Fisher Vector encoding (256 GMMs) and a
Hellinger kernel. Note that the resulting representation has 100k
dimension, \ie, is significantly higher dimensional. 
Furthermore,  STMH is an order of magnitude faster to extract.

\subsection{Comparison to the state of the art}

\begin{figure}
 \centering
 \resizebox{\linewidth}{!}{
 \begin{tabular}{m{0.75\linewidth}m{0.25\linewidth}}
  
  \hspace{-0.5cm}
 \includegraphics[width=1.1\linewidth,height=4.5cm]{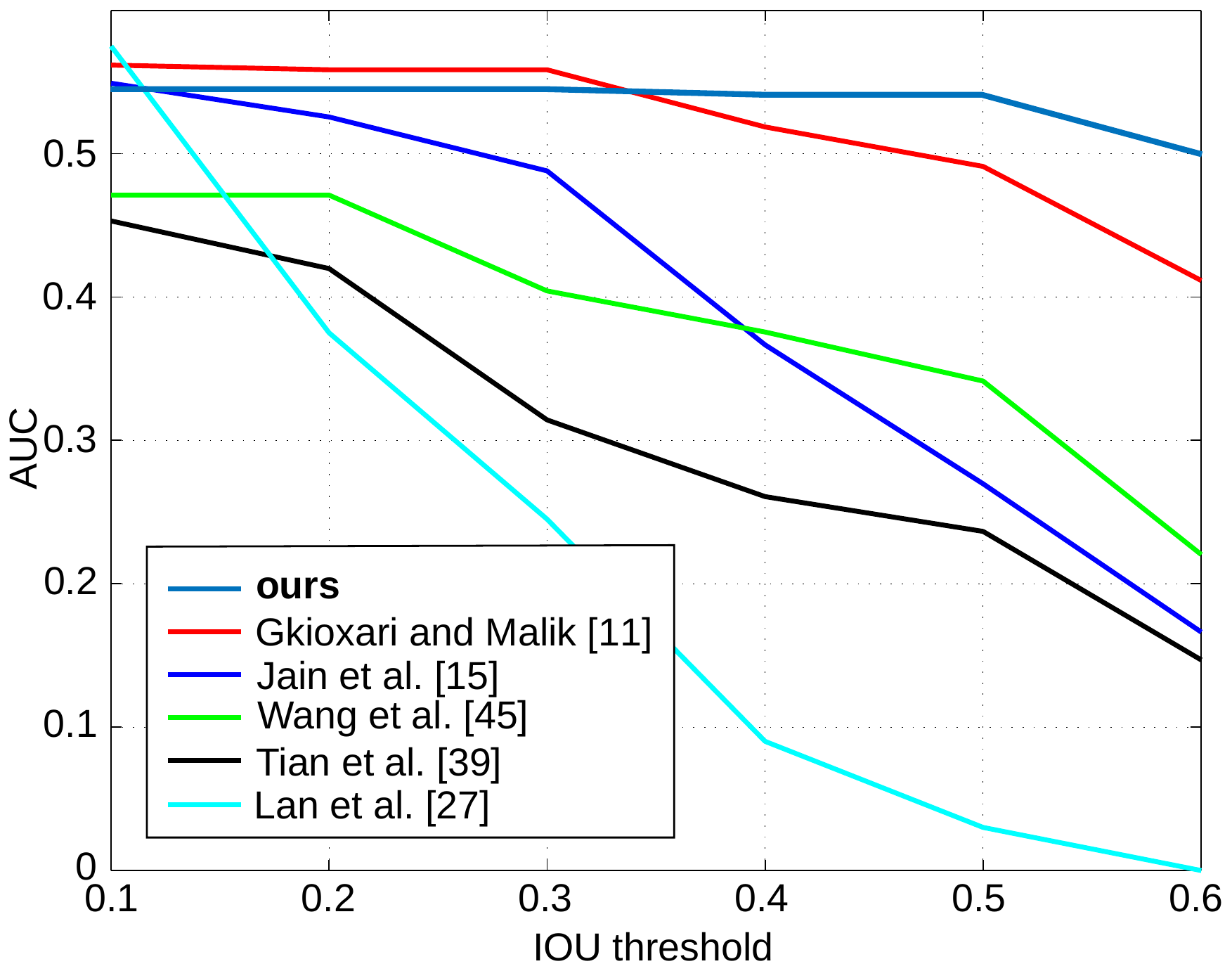} 
 & 
 \small
 \begin{tabular}{|c|c|} 
 \hline
 Method & mAP \\   
 \hline
 \cite{fat} & 75.8 \\
 \hline
 no STMH & 88.2 \\
 \textbf{ours} & \textbf{90.5} \\
 \hline
 \end{tabular} \\
 \end{tabular}}
  \normalsize
 \caption{Comparison to the state of the art on UCF-Sports. \textit{Left:} AUC for varying IoU thresholds. 
 \textit{Right:} mAP at $\delta = 50\%$. `no STMH' refers to our method without rescoring
 based on STMH.}
 \label{fig:ucfsports}
 
\vspace{-0.2cm}

\end{figure}

\begin{figure}
\centering
\hfill \includegraphics[width=0.32\linewidth]{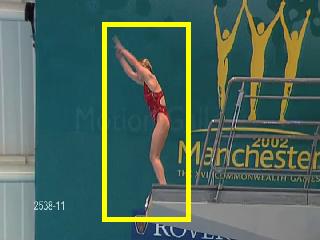} \hfill \includegraphics[width=0.32\linewidth]{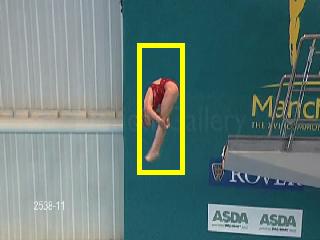} \hfill \includegraphics[width=0.32\linewidth]{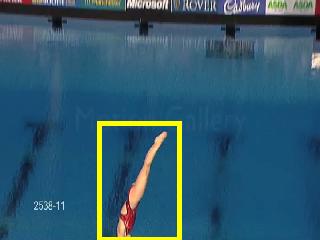} \hfill \\
\hfill \includegraphics[width=0.32\linewidth]{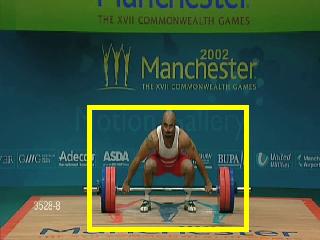} \hfill \includegraphics[width=0.32\linewidth]{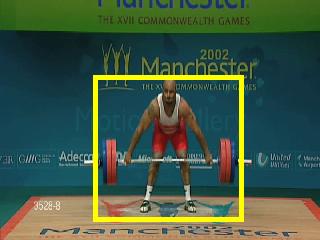} \hfill \includegraphics[width=0.32\linewidth]{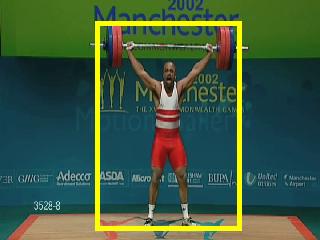} \hfill  \\
\caption{Example results from the UCF-Sports dataset.}
\label{fig:ex1}

\vspace{-0.4cm}

\end{figure}

\begin{figure*}
 \center
 \includegraphics[width=0.9\linewidth]{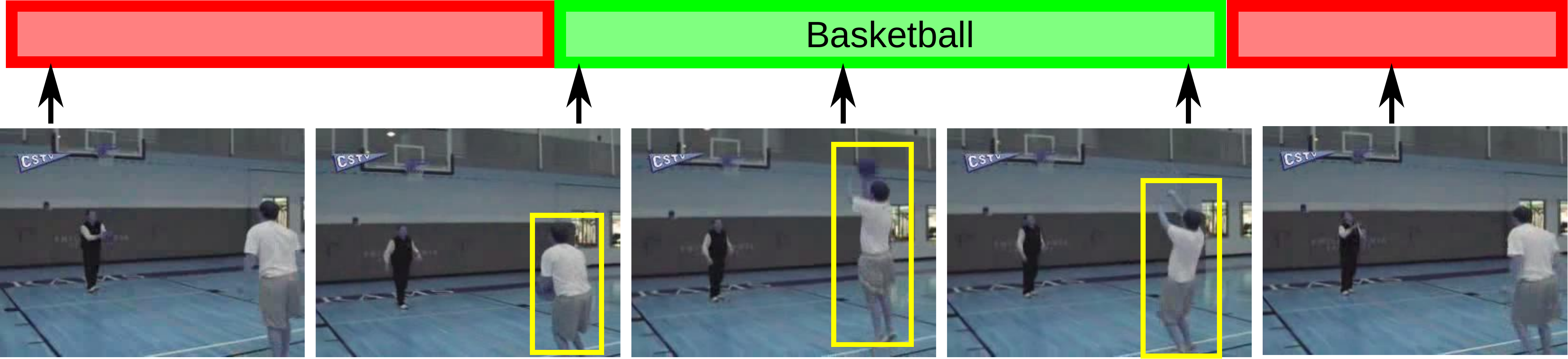}
 \caption{Example of spatio-temporal detection for the Basketball action on UCF-101.}
 \label{fig:extemp}
 
 \vspace{-0.35cm}
 
\end{figure*}

In this section, we compare our approach to the state of the art.
On UCF-Sports, past work usually give ROC curves and report Area Under the Curve (AUC). 
Figure~\ref{fig:ucfsports} (left) shows a comparison with the state of the art using the same protocol for different IoU thresholds $\delta$.
We can observe that our approach outperforms the state of the art. 
Note that at a low threshold, all methods obtain a comparable performance, but the gap widens for larger one, \ie, more precise 
detections.
Indeed, our spatial localization enjoys a high precision thanks to the tracking: the position of the detected region is refined in each frame using a sliding window.
As a consequence, the IoU between our detected tracks and the ground-truth is high, explaining why our performance remains constant between a low threshold and a high threshold $\delta$.
Figure~\ref{fig:ex1} shows example results. 
Despite important changes in appearance, the actor is successfully tracked throughout the video. 
For detection, mAP is more suitable as it does not depend on negatives.
Results are shown in Figure~\ref{fig:ucfsports} (right). 
We outperform the state of the art with a margin of $15\%$ and obtain a mAP of $90.5\%$. 
We also compute the mAP when scoring without STMH classifiers, \ie, the score is based on CNN features only, and observe a drop of $2\%$.

\begin{table}
 \centering
 
 \resizebox{\linewidth}{!}{
  \begin{tabular}{|c||c|c|c|c|}
   \hline
   $\delta$ & 0.2 & 0.3 & 0.4 & \textbf{0.5} \\
   \hline
   \cite{fat} & & & & 53.3 \\
   \hline
   no STMH & 58.1 \footnotesize{$\pm$ 2.1} & 58.0 \footnotesize{$\pm$ 1.9} & 57.7 \footnotesize{$\pm$ 2.1} & 56.5 \footnotesize{$\pm$ 2.6} \\
   \hline
   \textbf{ours} & 63.1 \footnotesize{$\pm$ 1.8} & 63.5 \footnotesize{$\pm$ 1.8} & 62.2 \footnotesize{$\pm$ 1.9} & 60.7 \footnotesize{$\pm$ 2.7} \\ 
   \hline
  \end{tabular}
}
  \caption{Comparison to the state of the art on J-HMDB using mAP for varying IoU thresholds $\delta$.
  We also report the standard deviation among the splits.}
  \label{tab:jhmdb}
  
\vspace{-0.25cm}
  
\end{table}

\begin{table}
 \centering
 \small
 \begin{tabular}{|c||c|c|c|c|}
  \hline
  $\delta$ & 0.05 & 0.1 & \textbf{0.2} & 0.3 \\
  \hline
  \cite{Yu_2015_CVPR} & 42.8 & & & \\
  \hline
  \textbf{ours} & 54.28 & 51.68 & 46.77 & 37.82\\
  \hline
 \end{tabular}

 \normalsize
 \caption{Localization results (mAP) on UCF-101 (split 1) for different IoU thresholds $\delta$.}
 \label{tab:ucf101}
 
 \vspace{-0.25cm}
 
\end{table}

The results for the J-HMDB dataset are given in Table~\ref{tab:jhmdb}.
We also outperform the state of the art by more than 7\% on J-HMDB at a standard threshold $\delta = 0.5$.
In particular, adding STMH leads to an improvement of 4\%.
We can also see that the mAP is stable \wrt the threshold $\delta$.
This highlights once again the high precision of the spatial detections, \ie, they all have a high overlap with the ground-truth, thanks to the tracking.

\begin{figure}
 \centering
\hfill \includegraphics[width=0.32\linewidth,height=1.6cm]{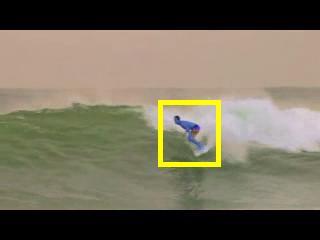} \hfill \includegraphics[width=0.32\linewidth,height=1.6cm]{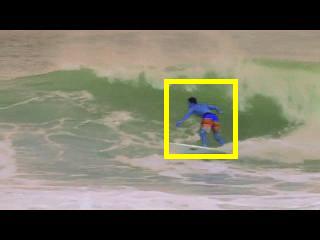} \hfill \includegraphics[width=0.32\linewidth,height=1.6cm]{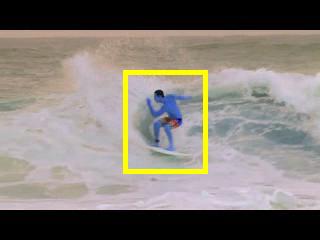}\hfill \\
\hfill \includegraphics[width=0.32\linewidth,height=1.8cm]{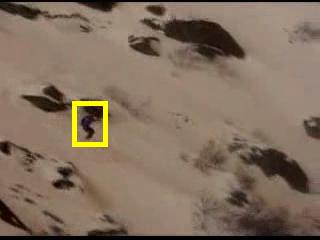} \hfill \includegraphics[width=0.32\linewidth,height=1.8cm]{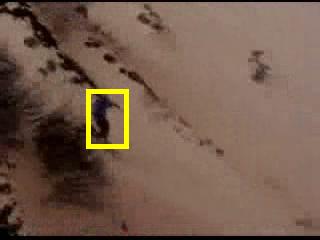} \hfill \includegraphics[width=0.32\linewidth,height=1.8cm]{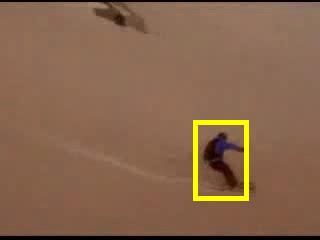}\hfill \\
\caption{Example results from the UCF-101 dataset.}
\label{fig:ex2}

\vspace{-0.5cm}

\end{figure}

Finally, we report the results for spatio-temporal detection on the UCF-101 dataset in Table~\ref{tab:ucf101}.
We obtain a mAP of more than $47\%$ at a standard threshold $\delta = 20\%$ despite the challenge of detecting an action both spatially and temporally.
At a threshold $\delta=5\%$, we obtain a mAP of $54\%$ compared to $42\%$ reported by~\cite{Yu_2015_CVPR}.
Figure~\ref{fig:extemp} and~\ref{fig:ex2} show example results. We can observe that  the result for the 
action ``Basketball'' is precise both in space and time. While most of
the 24 action classes cover almost the entire video, \ie, there is no
need for temporal localization, the action ``Basketball'' covers on
average one fourth of the video, \ie, it has the shortest relative
duration in UCF-101. For this class our temporal localization approach
improves the performance  significantly. The AP for Basketball is
$28.6\%$ ($\delta = 20\%$) with our full approach.
If we remove the temporal localization step, the
performance drops to $9.63\%$. This shows that our approach is capable
of localizing actions in untrimmed videos. 
With respect to tracking in untrimmed videos, tracking starts from the
highest scoring proposal in both directions (forward and backward) 
and continues even if the action is no longer present. The temporal
sliding window can then localize the action and removing 
parts without the action. Future work
includes designing datasets for spatio-temporal localization in
untrimmed videos in order to evaluate temporal localization more thoroughly.

\vspace{-0.2cm}

\section{Conclusion}
\label{sec:ccl}

\vspace{-0.1cm}

We present an effective approach for action localization, that detects actions
in space and time. 
Our approach builds upon object proposals extracted at the frame level that we track throughout the video.
Tracking is effective, as we combine instance-level and class-level detectors.
The resulting tracks are scored by combining classifiers learned on CNN features and our proposed spatio-temporal descriptors. 
A sliding window finally performs the temporal localization of the action.
The proposed approach improves on the state of the art
by a margin of 15\% in mAP on UCFSports, 7\% on J-HMDB and 12\% on UCF-101.

\vspace{0.1cm}

\paragraph{Acknowledgements} This work was partially supported by projects  ``Allegro'' (ERC),
 ``Khronos'' (ANR-11-LABX-0025), the MSR-Inria Joint Centre, a Google Faculty Research Award and the Moore-Sloan Data Science Environment at NYU.

\vspace{-0.2cm}

{\small
\bibliographystyle{ieee}
\bibliography{biblio}
}

\end{document}